\documentclass[sigconf]{acmart}

\usepackage{booktabs} 
\usepackage{float}
\usepackage{subfigure}
\usepackage{epstopdf}
\usepackage{array}
\usepackage{pifont}
\usepackage{multirow}
\usepackage{amssymb}
\usepackage{footnote}
\usepackage{tablefootnote}
\usepackage[section]{placeins}

\usepackage{soul}

\newcommand{\model}{\texttt{ZSCRGAN}}

\usepackage{color}
\newcommand{\todo}[1]{\textcolor{red}{TODO: #1}}
\newcommand{\new}[1]{\textcolor{blue}{#1}}

\usepackage{amsmath}
\usepackage{threeparttable}
\usepackage{algpseudocode,algorithm,algorithmicx}

\makeatletter
\def\BState{\State\hskip-\ALG@thistlm}
\makeatother
\newlength\myindent 
\AtBeginDocument{%
  \providecommand\BibTeX{{%
    \normalfont B\kern-0.5em{\scshape i\kern-0.25em b}\kern-0.8em\TeX}}}

\renewcommand\footnotetextcopyrightpermission[1]{}

\makeatletter
\def\@copyrightspace{\relax}
\makeatother

\begin{document}



\title{ZSCRGAN: A GAN-based Expectation Maximization Model for Zero-Shot Retrieval of Images from Textual Descriptions}
\titlenote{\textcolor{blue}{\bf This work has been accepted at the ACM Conference on Information and Knowledge Management (CIKM) 2020.}}

\author{Anurag Roy}
\affiliation{\institution{Department of CSE, IIT Kharagpur, India}}

\author{Vinay Kumar Verma}
\affiliation{\institution{Department of ECE, Duke University, USA}}

\author{Kripabandhu Ghosh}
\affiliation{\institution{Department of CSA, IISER Kolkata, India}}

\author{Saptarshi Ghosh}
\affiliation{\institution{Department of CSE, IIT Kharagpur, India}}

\renewcommand{\shortauthors}{A. Roy et al.}


\begin{abstract}
Most existing algorithms for cross-modal Information Retrieval are based on a supervised train-test setup, where a model learns to align the mode of the query (e.g., text) to the mode of the documents (e.g., images) from a given training set. 
Such a setup assumes that the training set contains an exhaustive representation of all possible classes of queries. 
In reality, a retrieval model may need to be deployed on previously unseen classes, which implies a \textit{zero-shot} IR setup.
In this paper, we propose a novel GAN-based model for zero-shot text to image retrieval. When given a textual description as the query, our model can retrieve relevant images in a zero-shot setup. 
The proposed model is trained using an Expectation-Maximization framework.
Experiments on multiple benchmark datasets show that our proposed model comfortably outperforms several state-of-the-art zero-shot text to image retrieval models, as well as zero-shot classification and hashing models suitably used for retrieval. 
\end{abstract}


\maketitle

\pagestyle{plain}  

\section{Introduction}

Today, information is generated in several modes, e.g., text, image, audio, video, etc. 
Thus, for a query in one mode (e.g., text), the relevant information may be present in a different mode (e.g., image). 
Cross-modal Information Retrieval (IR) algorithms are being developed to cater to such search requirements.

\vspace{2mm}

\noindent {\bf Need for Zero-shot Information Retrieval (ZSIR):}
A train-test setup of an IR task comprises parameter learning for various classes / categories of queries. Standard cross-modal retrieval methods require training data of all classes of queries to train the retrieval models. 
But such methods can fail to retrieve data for queries of {\it new} or {\it unseen} classes. 
For instance, suppose the retrieval model has been trained on images and textual descriptions of various classes of vehicles, such as `car', `motorbike', `aeroplane', and so on. Now, given a query `bus', the model is expected to retrieve images and textual descriptions of buses (for which the model has not been trained). 
Such a situation conforms to the ``zero-shot'' setup~\cite{Larochelle:2008:ZLN:1620163.1620172,ZeroShotNIPS2009,ZeroShotNorouzi2013}  which focuses on recognizing new/unseen classes with limited training classes.

Such situations are relevant in any modern-day search system, where new events,  hashtags, etc. emerge every day. 
So, contrary to the conventional IR evaluation setup, the zero-shot paradigm needs to be incorporated in an IR setting.
Specifically, zero-shot cross-media retrieval intends to achieve retrieval across multiple modes (e.g., images to be retrieved in response to a textual query) where there is no overlap between the query-classes in training and test data. 
Zero-Shot IR (ZSIR) is especially challenging since models need to handle not only different semantics across seen and unseen query-classes, but also the heterogeneous features of data across different modes.

\vspace{2mm}
\noindent {\bf Present work and differences with prior works:}
Though lot of research has been reported on general multimodal and cross-modal retrieval~\cite{cross-modal-ir-survey}, to our knowledge, only a few prior works have attempted cross-modal IR in a zero-shot setting~\cite{scottreed-cvpr16,ZSL_GAN_cvpr2018,ijcai2018-92,8643797,AgNet-TNNLS2019, CZHash-ICDM2020} (see Section~\ref{sec:related} and Section~\ref{sub:baselines} for details of these methods).
Some of these prior works assume additional information about the class labels (e.g., a measure of semantic similarity between class labels) whcih may not always be available. 

In this paper, we propose a novel model for cross-modal IR in zero-shot setting, based on Conditional Generative Adversarial Networks (GANs)~\cite{mirza-conditional14}, that can retrieve {\it images} relevant to a given {\it textual} query.
Our model -- which we name {\bf \model{}} (Zero-Shot Cross-modal Retrieval GAN) -- relies only on the textual data to perform the retrieval, and {\it does not need additional information about class labels}. 
Though prior ZSIR models~\cite{ZSL_GAN_cvpr2018,ijcai2018-92,8643797} also use GANs, the main novel contributions of the proposed model can be summarized as follows:
(1)~We propose \textbf{use of wrong classes} to enable the generator to generate features that are unique to a specific class, by distinguishing it from other classes.
(2)~We develop a \textbf{Common Space Embedding Mapper (CSEM)} to map both the image embeddings and the text embeddings to a common space where retrieval can be performed. {\it This is the key step that enables our model to perform retrieval without relying on additional semantic information of class labels.}
(3)~We develop an {\bf Expectation-Maximization (E-M) based method} for efficiently training the retrieval model, where the GAN and the CSEM are trained alternately. We show that this E-M setup enables better retrieval than jointly training the GAN and the CSEM.

We experiment on several benchmark datasets for zero-shot retrieval -- (1)~the Caltech-UCSD Birds dataset, (2)~the Oxford Flowers-102 dataset, (3)~the North America Birds (NAB) dataset, (4)~the Wikipedia dataset, and (5)~the Animals With Attribute (AWA) dataset. 
Our proposed \model{} comfortably out-performs several strong and varied baselines on all the datasets, including ZSIR models~\cite{scottreed-cvpr16,8643797,ZSL_GAN_cvpr2018}, state-of-the-art Zero-Shot classification models~\cite{verma-cvpr18,xian2018feature_gen} suitably adapted for the Text-to-Image retrieval setting, 
as well as state-of-the-art Zero-Shot Hashing models~\cite{AgNet-TNNLS2019, CZHash-ICDM2020,DCMH-AAAI17}. 
Also note that our proposed model can be used not only with textual queries, but also with other forms of queries such as attribute vectors, as demonstrated by its application on the AWA dataset. We make the implementation of \model{} publicly available at \url{https://github.com/ranarag/ZSCRGAN}.

\section{Related work} \label{sec:related}

There are many works on multimodal retrieval (see~\cite{cross-modal-ir-survey} for a survey). 
However, most of these works are {\it not} in zero-shot setup on which we focus in this paper.

\vspace{1mm}
\noindent {\bf Zero-Shot Learning (ZSL):}
The initial models for ZSL mostly focused on learning a similarity metric in the joint attribute space or feature space~\cite{dem-zsl,cmt-zsl}. 
With the recent advancements of the generative model~\cite{vae-kingma,gan-nips14}, models based on Variational Autoencoders (VAE)~\cite{verma-cvpr18} and GAN-based~\cite{xian2018feature_gen} approaches have attained the state-of-the-art results for ZSL.

\vspace{2mm}
\noindent {\bf Multimodal IR in Zero-Shot setup (ZSIR):}
There have been several recent works on multimodal ZSIR. 
For instance, some works have attempted zero-shot {\it sketch-based} image retrieval~\cite{Dutta2019SEMPCYC, Dey_2019_CVPR, SBIR_2018_ECCV}, most of which use GANs to perform the retrieval. 
Note that sketch-based image retrieval is different from the text-to-image retrieval that we consider in this paper -- while the input query is a sketch in the former, we assume the query to be a textual description (or its vector representation).

There have also been works on zero-shot text-to-image retrieval. 
Reed {\it et al.}~\cite{scottreed-cvpr16} minimized empirical risk function to create a joint embedding for both text and images.
Chi {\it et al.} proposed two very similar models~\cite{8643797} and~\cite{ijcai2018-92}.
The DADN model developed in~\cite{8643797} was shown to out-perform the one in~\cite{ijcai2018-92}.
These models adopt a dual GAN approach where a forward GAN is used to project the embeddings to a semantic space, and a reverse GAN is used to reconstruct the semantic space embeddings to the original embeddings. 
Zhu {\it et al.}~\cite{ZSL_GAN_cvpr2018} developed a GAN-based model named ZSL-GAN for retrieving images from textual queries, where the GAN is trained with classification loss as a regularizer.
Additionally, several Zero-Shot Hashing models have been developed~\cite{AgNet-TNNLS2019, CZHash-ICDM2020,DCMH-AAAI17,SePH-ITC17} that can also be used for ZS text-to-image retrieval. 

\vspace{1mm}
\noindent All the above-mentioned prior models for text-to-image ZSIR are considered as baselines in this work.
Further details of the prior models are given in Section~\ref{sec:expt}, where the primary differences of these models with our proposed model are also explained.


\section{Proposed Approach}

We start by formally defining the zero-shot text to image retrieval problem, and then describe our proposed model \model{} (implementation available at \url{https://github.com/ranarag/ZSCRGAN}). 
Table~\ref{tab:notations} gives the notations used in this paper.

\begin{table}[t]
\scriptsize
	\centering
	\addtolength{\tabcolsep}{-1pt}
	\begin{tabular}{|l|p{0.35\columnwidth}||l|p{0.35\columnwidth}|}
		\hline
			& \textbf{Description} & 	& \textbf{Description}\\ 
		\hline
		$G()$ &Generator & $D()$ &Discriminator\\
		\hline	
		${L}_{D}$ & Discriminator Loss Function & $\mathcal{L}_{G}$ & Generator Loss Function \\
		\hline
		$\varphi_{t}$ & Text embedding of unseen class & $I_i$ & Image embedding of unseen class\\ 		
		\hline
		$\theta_{t}$ & Common Space embedding generated from $\varphi_{t}$ & $\theta_{i}$ & Common space embedding generated from $I_i$ \\
		\hline

		$I_{r}$ & Real Image Embedding & $I_{w}$ & Wrong Image Embedding \\
		\hline
		$i_{r}$ & Real representative Embedding & $i_{w}$ & Wrong representative Embedding \\
		\hline
		$\varphi_{t_r}$ & Real Text Embedding & $\varphi_{t_w}$ & Wrong Text Embedding \\
		\hline
		$\hat{c_{tr}}$ & Real Latent Embedding & $\hat{c_{tw}}$ & Wrong Latent Embedding \\
		\hline
		$\psi_{C}$ & trainable parameters of CSEM & $\psi_{G}$ & trainable parameters of Generator \\
		\hline
		
		$\theta_{tr}$ & Common Space embedding generated from $\varphi_{tr}$ & $\theta_{tw}$ & Common space embedding generated from $\varphi_{tw}$ \\
		\hline
		$\psi_{C}$ & Trainable parameters of CSEM & $\psi_{G}$ & Trainable parameters of Generator \\
		\hline
		$\hat{c_{t}}$ & Latent Embedding Generated from $\varphi_{t}$ & z & noise vector sampled from a normal distribution $p_z$ \\
		\hline
	\end{tabular}
	\caption{{\bf Notations used in this paper.}}
    \label{tab:notations}
    \vspace{-10mm}
  \end{table}

\vspace{-3mm}
\subsection{Problem Definition}

In the zero-shot Text to Image retrieval setup, we consider training data $\mathcal{D}=\{I_r^k,\varphi_{t_r}^k,y^k\}_{k=1}^K$ with $K$ samples. 
$I_r^k$ is the real image embedding of the $k^{th}$ image. 
$\varphi_{t_r}^k$ is the real text embedding of the text accompanying the $k^{th}$ image.
$y^k \in \mathcal{Y}$ is a class label, and $\mathcal{Y}=\{1,2,\dots,S\}$ is the set of {\it seen} classes, where $S$ is the number of seen classes (only seen classes are available at training time). 

Let $\mathcal{U}=\{1,2,\dots,U\}$ be the set of unseen classes, where $U$ is the number of unseen classes (not available at training time). For each unseen class query $u \in \mathcal{U}$ a relevant set of images are present that we have to retrieve.
At test-time, for each unseen class $u \in \mathcal{U}$, we use a textual embedding $\varphi_{t}$ from $u$ as query. Textual embedding $\varphi_{t}$ and unseen class image are projected into joint space to perform the retrieval. In the zero-shot setup $\mathcal{U} \cap \mathcal{Y}=\Phi$, i.e. training and test classes are disjoint. 



\begin{figure*}[!t]
\centering
\includegraphics[height=5cm,width=14cm]{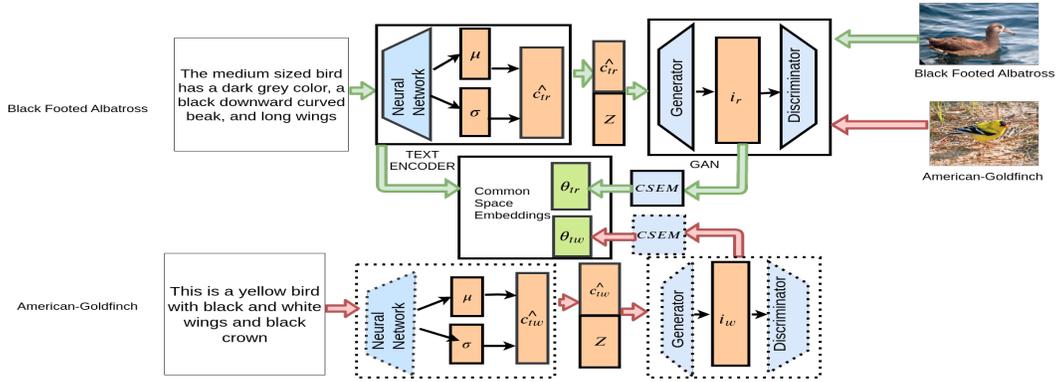}
\caption{{\bf [color online] The proposed \model{} architecture used for generating a common space embedding for text and image. The example here demonstrates training of the architecture for learning a common space embedding for the class `Black Footed Albatross', using textual descriptions and images of that class. 
To this end, textual descriptions and images of some {\it other} class (here `American-Goldfinch') are used. We refer to this other class as the `wrong' class. 
The blocks in dashed lines (through which textual embeddings of the wrong class are passed) are identical copies of the corresponding blocks in solid line (through which textual embeddings of the correct class are passed).}}
\label{fig:embedding_model}
\vspace{-5mm}
\end{figure*}

\subsection{Overview of our approach}
One of the main challenges of zero-shot cross-modal retrieval is the \textit{representation problem} of \emph{novel (unseen) class} data. The text and image are of different modalities and the challenge is to represent them in a common embedding space. We train a model to perform this mapping. However training such a model becomes difficult when there is a huge overlap among embeddings of different classes, e.g., there may be cases where the images of one class have high overlap with images from other classes, making it difficult to map it to a common embedding space.
As part of addressing this challenge, we use a Generative Adversarial Network (GAN)~\cite{gan-nips14} to generate a per class `representative embedding'  for all image embeddings of a particular class.
This generated embedding is desired to have {\it high} similarity with image embeddings from the same class and {\it low} similarity with image embeddings from other classes.

We chose a generative model rather than a discriminative one for this purpose, because generative models are most robust to visual imagination, and this helps the model to map the images from unseen classes more robustly.
Two popular generative models are Variational Autoencoders (VAE)~\cite{vae-kingma} and Generative Adversarial Networks (GAN)~\cite{gan-nips14}. 
Samples from VAE models tend to be blurry (i.e., with less information content) as compared to GANs, because of the probability mass diffusion over the data space~\cite{Theis2015d}. 
Therefore, we found GANs to be the most suitable for this problem. 

While there have been prior works using GANs for zero-shot retrieval tasks~\cite{8643797,ijcai2018-92,scottreed-cvpr16}, they rely on class labels for training. Some of the prior models~\cite{8643797,ijcai2018-92} make use of word-embeddings of class labels as class level information. 
However, in many cases, class label information may not be available.
In contrast, our proposed model does {\it not need} class labels to perform its task. This is achieved by learning to map image embeddings and text embeddings to a common embedding space.

Our proposed model \model{} (shown in Figure~\ref{fig:embedding_model}) works as follows.
We take two text embeddings, one belonging to class $y \in \mathcal{Y}$ and the other belonging to some other class $\hat{y} \in \mathcal{Y}$. We pass the two text embeddings through a Text Encoder (TE) which generates 
(i)~a latent embedding $\hat{c_{tr}}$ (a.k.a real text embedding) for class $y$ and (ii)~$\hat{c_{tw}}$ (a.k.a wrong text embedding) for class $\hat{y}$. 
For each class, we generate a per-class representative embedding $i$ for all images of that class. We train the Text Encoder jointly with the GAN. We use the representative embeddings to train a {\it Common Space Embedding Mapper} (CSEM) which learns to map all the image and text embeddings to a common space where these common space embeddings will have a high similarity if they belong to the same class.

We formulate an E-M paradigm in which we train the CSEM and the GAN alternatively. In a latter section we also justify our choice of such a paradigm by comparing the model performance with this E-M formulation and without it (i.e., on jointly training the CSEM and GAN).

\subsection{Details of our proposed model \model{}} \label{sec:EMsetup}
Let $Q(I, \phi)$ be the  joint probability distribution denoting the probability of text embeddings $\phi$ and relevant image embeddings $I$.
Maximizing this probability is expected to ensure high similarity between $\phi$ and relevant image embeddings $I$ therefore leading to better performance of the retrieval model. 
We plan to do this maximization using a machine learning model having $\psi_{C}$ as its parameters. 
Hence, our aim will be to maximize the probability distribution $Q(I, \phi | \psi_{C})$.
For simplicity, we will maximize the log of this distribution $\log Q(I, \phi| \psi_{C})$. 
.Let $\psi_{C}$ be the random variable representing the values that can be taken by the trainable parameters of the neural network model. Thus, our log probability function becomes $\log Q(I, \phi | \psi_{C})$. 
However, when we try to maximize $\log Q(I, \phi | \psi_{C})$ directly using a  machine learning model , we see very poor retrieval performance. The reason being, images from one class are very similar to images from another class . For example, {\it `Crested Auklet'} and {\it`Rhinoceros Auklet'} in the CUB dataset are two very similar looking birds and are almost indistinguishable to the human eye. Due to these overlaps among images from classes, training the neural network model becomes difficult, as it encounters very similar positive and negative examples during training. Thus , we introduce 
a latent variable $I'$ -- an unobserved embedding which would be a representative for the images of a class. The embedding will have high similarity with the images of a particular class and very low similarity with images from all other classes. Using these representative embeddings instead of the actual image embeddings will solve the training problem of  our model . 
We adopt an Expectation-Maximization (E-M) framework to perform the maximization in presence of the latent variable.

\subsubsection{{\bf E-M formulation:}}

As stated above, our objective is to maximize the following expression:
\begin{equation}
    \small
        \log Q(I, \phi | \psi_{C}) \; = \log  \sum_{i \in I'} \; Q(i, I, \phi | \psi_{C})\; 
\end{equation}
Let $P(I' = i | I, \phi, \psi_{G})$ denote the probability of generating $i$ given $I$, $\phi$ and $\psi_{G}$. Here is $\psi_{G}$ are the trainable parameters of the  model  used to generate $i$. Thus we have:
\begin{equation}
\small
\label{eqn:em-derivation}
    \begin{aligned}
        \log & \sum_{i \in I'}   Q(i, I, \phi | \psi_{C}) \; =    \log \sum_{i \in I'} \frac{ P(i | I, \phi, \psi_{G}) \cdot Q (i, I, \phi | \psi_{C})}{P(i | I, \phi, \psi_{G})} \\
        &\; \geq  \sum_{i \in I'} P(i | I, \phi, \psi_{G}) \cdot \log \frac{Q(i, I, \phi | \psi_{C})}{P(i | I, \phi, \psi_{G})} \;   \; \mbox{by Jensen's Inequality}\\
        & = \sum_{i \in I'} P(i|I, \phi, \psi_{G}) \cdot \log Q(i, I, \phi | \psi_{C}) \; - \; \sum_{i \in I'} P(i|I, \phi, \psi_{G}) \cdot \log P(i| I, \phi, \psi_{G}) \\
        & = \sum_{i \in I'} P(i|I, \phi, \psi_{G}) \cdot [\log Q(i,\phi | I, \psi_{C}) + \log Q(I | \psi_{C})] \\
        & \;\;\;\;\; - \; \sum_{i \in I'} P(i|I, \phi, \psi_{G}) \cdot \log P(i| I, \phi, \psi_{G}) \\
        & \mbox{since $I$ is conditionally independent of $i$ and $\phi$ given $\psi_{C}$} \\
        & = \sum_{i \in I'} P(i|I, \phi, \psi_{G}) \cdot \log Q(i,\phi | \psi_{C})\; +\; \sum_{i \in I'} P(i|I, \phi, \psi_{G}) \cdot\log Q(I | \psi_{C}) \\ 
        & \;\;\;\;- \; \sum_{i \in I'} P(i|I, \phi, \psi_{G}) \cdot \log P(i| I, \phi, \psi_{G}) \\
        & = \sum_{i \in I'} P(i|I, \phi, \psi_{G}) \cdot \log Q(i,\phi | \psi_{C})\; + \;1 \cdot \log Q(I) \\
        & \;\;\;\;\; - \; \sum_{i \in I'} P(i|I, \phi, \psi_{G}) \cdot \log P(i| I, \phi, \psi_{G})
        \end{aligned}
\end{equation}
where the last step holds since the distribution of $I$ is independent of $\psi_{C}$.
Now, after applying Jensen's Inequality Eqn.~\ref{eqn:em-derivation} implies the following:
\begin{equation}
\small
\label{eqn:em-derivation-2}
\begin{aligned}
 \log & \sum_{i \in I'}  Q(i, I, \phi | \psi_{C}) -  \log Q(I) \\
        & \geq \sum_{i \in I'} P(i|I, \phi, \psi_{G}) \cdot \log Q(i,\phi | \psi_{C})\; - \; \sum_{i \in I'} P(i|I, \phi, \psi_{G}) \cdot \log P(i| I, \phi, \psi_{G}) \\
        & = F(P, \psi_{C})
    \end{aligned} 
\end{equation}
$Q(i,\phi | \psi_{C})$ denotes the joint probability of $i$ and $\phi$ being similar.
Thus, $F(P, \psi_{C})$ is a lower bound for  $\log Q(I, \phi | \psi_{C})$, and maximizing the lower bound will ensure a high minimum value of $\log Q(I, \phi | \psi_{C})$.\footnote{The function $F(P, \psi_{C})$ is related to the Free Energy Principle that is used to bound the `surprise' on sampling some data, given a generative model (see~\url{https://en.wikipedia.org/wiki/Free_energy_principle}).} 
We train a neural network architecture using the E-M algorithm to maximize $F(P,\psi_{C})$ where the E and M steps at iteration \textbf{$it$} are:
\begin{align}
        \mbox{E-step} &: \; P^{it} (I'=i|I, \phi) = \underset{P}{\operatorname{argmax}} \; F(P, \psi_{C}^{it-1}) \\
        \mbox{M-step} &: \; \psi_{C}^{it} = \underset{\psi_{C}}{\operatorname{argmax}} \; F(P^{it}, \psi_{C})        
\end{align}
So, the challenge now is to maximize $F(P, \psi_{C})$. To this end, we propose to use neural networks as follows.

\subsubsection{{\bf Using neural networks to approximate $F(P, \psi_{C})$:}} 
 Eqn.~\ref{eqn:em-derivation} can be re-written as:
\begin{equation}
\label{eqn:F}    
    \begin{aligned}
    F(P, \psi_{C}) &\; = \mathbb{E}_{i \sim P(I'=i| I, \phi, \psi_{G})}[\log Q(i, \phi | \psi_{C})] \; \\
                 &\; - \mathbb{E}_{i \sim P(I'=i| I, \phi, \psi_{G})}[\log P(i | I, \phi, \psi_{G})] \; \\
    \end{aligned}
\end{equation}
We take the help of two neural networks to approximate the two parts of the function $F$ as shown in Eqn.~\ref{eqn:F} -- (1)~Common Space Embedding Mapper (CSEM) to represent the first term, and (2)~GAN to represent the second term in Eqn.~\ref{eqn:F}.

\vspace{2mm}
\noindent {\bf Common Space Embedding Mapper (CSEM):} This module is a feed-forward neural network trained to maximize the probability  $Q(i, \phi | \psi_{C})$ which denotes the joint probability of $i$ and $\phi$ being similar.
We define such a $Q$ as:
\begin{equation}
    \small
    \begin{aligned}
    Q(i, \phi | \psi_{C}) &\; = \frac{e^{v_p}}{e^{v_p} + e^{v_n}} \; \\
    \end{aligned}
\end{equation}

where $v_p$ and $v_n$ are scores calculated as:
\begin{equation}
    \small
    \begin{aligned}
    v_p &\; =  CosineSim(CSEM(i), \hat{c_{tr}})\; \\
    v_n &\; =  CosineSim(CSEM(G(z,\hat{c_{tw}})), \hat{c_{tr}})\; \\
    \end{aligned}
\end{equation}
The CSEM is trained using the cost function Triplet Loss~\cite{Dong_2018_ECCV} $\mathcal{L}_{T}$, which can be written as:
\begin{equation}
\label{eq:TripletLoss}
\small
\begin{aligned}
\mathcal{L}_{T} &\; = -\mathbb{E}_{(i \sim P(I'=i | I, \phi, \psi_{G}), \hat{c_{tr}} \sim TE(\phi)} [\log Q(i, \phi | \psi_{C})] \; \\
&\; = \mathbb{E}_{(i \sim P(I'=i | I, \phi, \psi_{G}), ( \hat{c_{tr}},\hat{c_{tw}}) \sim TE(\phi)} [-\log \frac{e^{v_p}}{e^{v_p} + e^{v_n}}] \; \\
&\; = \mathbb{E}_{(i \sim P(I'=i | I, \phi, \psi_{G}), ( \hat{c_{tr}},\hat{c_{tw}}) \sim TE(\phi)} [\log (1 + e^{v_n - v_p})] \; \\
\end{aligned}
\end{equation}
We call this module the {\it Common Space Embedding Mapper} because it learns to map the image embeddings to a space where the resulting embeddings will have high cosine similarity among themselves if they are from the same class and low cosine similarity with images from different classes. We train it by using the triplet loss $\mathcal{L}_{T}$ considering $\hat{c_{tr}}$ as the pivot, $i$ as the positive example and $G(z,\hat{c_{tw}})$ as the negative example. Here, $\hat{c_{tr}}$ and $\hat{c_{tw}}$ are generated by $TE(\phi_{tr})$ and $TE(\phi_{tw})$ respectively, $TE()$ being the Text Encoder (described in Section~\ref{TextEncoder}).

\vspace{2mm}
\noindent {\bf GAN based Learning:} 
$\mathbb{E}_{i \sim P(I'=i | I, \phi)}[\log P(i | I, \phi)]$ is calculated using a Generative Adversarial Network.
Generative adversarial networks (GAN)~\cite{gan-nips14} are composed of two neural network models -- the {\it Generator} and the {\it Discriminator}, which  are trained to compete with each other. The generator (G) is trained to generate samples from a data distribution, $p_{g}$ that are difficult to distinguish from real samples for the discriminator. Meanwhile, the discriminator (D) is trained to differentiate between real samples sampled from the true data distribution $p_{data}$ and images generated by G. The GAN training is often very unstable; To stabilize the GAN training, Arjovsky et al.~\cite{pmlr-v70-arjovsky17a} proposed Wasserstein-GAN (WGAN).  
WGAN optimizes the Wasserstein (a.k.a Earth Mover's) distance between $p_{g}$ and $p_{data}$. The loss function to optimize in WGAN is:
\begin{equation}\label{eq:GAN_WD}
\small
\begin{aligned}
    \min_{G} \max_{D} V(D,G) = &\; \mathbb{E}_{x \sim {p_{data}}, z \sim {p_{g}}} [D(x) -  D(G(z)))] \; \\
&\; s.t. \lVert D_w \rVert \leq k
\end{aligned}
\end{equation}
where $D_w$ indicates all the parameters of the discriminator and $k$ is a constant.
\par
The WGAN architecture does not have control over the data generation of a particular class, which is a requirement in our application. Hence, to enable generation of data of a particular class, we use the \emph{Conditional-WGAN}~\cite{mirza-conditional14} (CWGAN), where the WGAN is conditioned on some latent embedding.
The proposed model uses the CWGAN as the base architecture, conditioned on the latent text embedding (which helps to achieve robust sample generation).
\par
The discriminator and generator losses are as follows:
\begin{equation}\label{eq:D_new}
\small
\begin{aligned}
\mathcal{L}_{D} &\; =  \frac{1}{2}(\ \mathbb{E}_{(I_{w},\varphi_{t_r}) \sim {p_{data}}} [ D(I_{w}, \varphi_{t_r})] \;  \\
+ &\;  \mathbb{E}_{z \sim {p_{z}}, \varphi_{t_r} \sim {p_{data}},  \hat{c_{tr}} \sim TE(\phi)} [D(G(z,\hat{c_{tr}}), \varphi_{t_r})] \;)  \\
- &\; \mathbb{E}_{(I_{r},\varphi_{t_r}) \sim {p_{data}}} [D(I_{r}, \varphi_{t_r})]
\end{aligned}
\end{equation}
\begin{equation}\label{eq:G_new}
\small
\begin{aligned}
\mathcal{L}_{G} & = -\mathbb{E}_{z \sim {p_{z}}, \varphi_{t_r} \sim p_{data}, \hat{c_{tr}} \sim TE(\phi)} [D(G(z,\hat{c_{tr}}), \varphi_{t_r})] \; \\
                    + &\; \alpha * (D_{JS}(\mathcal{N}(\mu(\varphi_{t_r}), \Sigma(\varphi_{t_r})) \, || \, \mathcal{N}(0, I) \; \\
                    + &\;  D_{JS}(\mathcal{N}(\mu(\varphi_{t_w}), \Sigma(\varphi_{t_w})) \, || \, \mathcal{N}(0, I)) \; \\
                    + &\; \beta * -\mathbb{E}_{i \sim P(I'=i| I, \phi, \psi_{G}), (I_{r}, I_{w}) \sim p_{data}}[ R(I'=i, I)] \; \\
\end{aligned}
\end{equation}
where $I_{r}$ and $\varphi_{t_r}$ are the image and text embeddings that belong to the same class.
Any image embedding {\it not} belonging to this corresponding class will be a candidate wrong image embedding $I_w$ and any text embedding not belonging to this class will be a candidate wrong text embedding $\varphi_{tw}$.
The latent embeddings $\hat{c_{tr}}$ and $\hat{c_{tw}}$ are sampled from the Normal Distribution with $\mathcal{N}(\mu(\varphi_{t}), \Sigma(\varphi_{t}))$, that is the latent space representation of the text embedding. This allow a small perturbations in $\varphi_{tr}$ and $\varphi_{tw}$ which is required to increase the generalizability of the model.
The function $R(I'=i, I)$ is a regularizer in order to ensure that the embeddings generated by the generator can be used to retrieve all the relevant images. The equations are as follows:
\begin{equation}\label{eq:R_new}
\small
\begin{aligned}
R(I'=i, I) &\; =- \log \frac{p(I'=i, I)}{1-p(I'=i, I)} \; \\
\end{aligned}
\end{equation}
where $p$ denotes the joint probability of $i$ and $I$ which we define as follows:
\begin{equation}
p(i, I)  = \frac{e^{-d_p}}{e^{-d_p} + e^{-d_n}}
\end{equation}
where $d_p = | I_{r} - i |$ and $d_n = | I_{w} - i | - \lambda$  are the Manhattan distances between the generated embedding $i$ and $I_r$ and $I_w$ respectively. We have also provided a margin $\lambda$ in order to ensure that $i$ is atleast $\lambda$ separated from $I_w$ in terms of Manhattan distance. 
Our formulation of probability $p(i, I)$ also ensures that  $p \in (0,1)$, since for $p$ to attain the value $0$ or $1$, $d_p$ or $d_n$ will have to be tending to $\infty$. However, in our formulation, $d_p$ and $d_n$ always attain finite values.
This ensures that the term $R(I'=i, I)$ is not undefined anywhere during our optimization. The log of odds ratio function can be further simplified as:
\begin{equation}
\log \frac{p(i, I)}{1-p(i, I)} = \log \frac{\frac{e^{-d_p}}{e^{-d_p} + e^{-d_n}}}{\frac{e^{-d_n}}{e^{-d_p} + e^{-d_n}}} 
= \log \frac{e^{-d_p}}{e^{-d_n}} 
= d_n - d_p 
\end{equation}



Here the discriminator ($D$) and generator ($G$) are trained by alternatively maximizing $L_D$ and minimizing $L_G$. Also, $\alpha$ and $\beta$, are hyper-parameters to be tuned to achieve optimal results.

\vspace{2mm}
\noindent {\bf Discriminator:}
In this adversarial game between the $D()$ and the $G()$, the $D()$ is trained to separate a real image from a fake one. In our task of  generating representative embedding $i \in I'$ from text embeddings of a particular class, an image embedding from a different class $I_w$ should also be identified by the discriminator as fake given the text embedding $\hat{c_{tr}}$. For example as in Figure~\ref{fig:embedding_model}, if the text embedding of \textit{Black-Footed-Albatross} is given and $G()$ generates an embedding of any other class, say \textit{American Goldfinch}, then $D()$ should label it as fake and force $G()$ to  generate $i$ for Black-Footed-Albatross from text-embedding of Black-Footed-Albatross. 
Hence we added another condition to the discriminator loss of classifying the image $I_w$ as fake and gave it equal weightage as compared to discriminator loss of classifying the generated  embedding as fake. Hence, as shown in Eqn.~\ref{eq:D}, 
we add another discriminator loss to the CWGAN discriminator loss:
\begin{equation}\label{eq:D}    
\small
\begin{aligned}
\mathcal{L}_{D} &\; =  (\mathbb{E}_{z \sim {p_{z}}, \varphi_{t_r} \sim p_{data}, \hat{c_{tr}} \sim TE(\phi)} [D(G(z,\hat{c_{tr}}), \varphi_{t_r})] \\
                &\; -  \mathbb{E}_{(I_{r},\varphi_{t_r}) \sim {p_{data}}} [D(I_{r}, \varphi_{t_r})])* 0.5 \;  \\
                &\; +  (\mathbb{E}_{(I_{w},\varphi_{t_r}) \sim {p_{data}}} [D(I_{w}, \varphi_{t_r})] \; \\ 
                &\; -  \mathbb{E}_{(I_{r},\varphi_{t_r}) \sim {p_{data}}} [D(I_{r}, \varphi_{t_r})] ) * 0.5 \; \\
\end{aligned}
\end{equation}

\vspace{2mm}
\noindent {\bf Generator:}
As shown in Figure \ref{fig:embedding_model}, we design a  $G$ (with the loss function stated in Eqn.~\ref{eq:G_new}) which takes the latent embedding vector as input and tries to generate image embedding of the same class. 
In order to achieve this goal, we add two regularizers to $G$ -- 
(1)~Negative Log of odds regularizer (Eqn.~\ref{eq:R_new}),(2)~Jensen-Shannon Divergence (Eqn.~\ref{eq:GaussianJS}). 
The intention for jointly learning the text encoder model instead of training the text encoder separately is to ensure that the text encoder model learns to keep the important features required for the generator to generate image embedding.


\vspace{2mm}
\noindent {\bf{Learning text encodings:}}\label{TextEncoder}
As shown in Figure~\ref{fig:embedding_model}, the text embedding $\varphi_{t}$ is first fed into an encoder which encodes $\varphi_{t}$ into a Gaussian distribution $\mathcal{N}(\mu(\varphi_{t}), \Sigma(\varphi_{t}))$, where mean $\mu$ and standard deviation $\Sigma$ are functions of $\varphi_{t}$. 
This helps to increase the samples of the text encoding and hence reduces the chance of overfitting.  
This kind of encoding provides small perturbations to  $\varphi_{t}$, thus yielding more train pairs given a small number of image-text pairs. We train this model by optimizing the following condition as a regularization term while training the generator.
\vspace{-5pt}
\begin{equation}\label{eq:GaussianJS}
D_{JS}(\mathcal{N}(\mu(\varphi_{t}), \Sigma(\varphi_{t})) \, || \, \mathcal{N}(0, I))
\vspace{-1pt}
\end{equation}
where $D_{JS}$ is the Jensen-Shanon divergence (JS divergence) between the conditioning distribution and the standard Gaussian distribution. Unlike the previous conditional GANs~\cite{mirza-conditional14} our model does not append $z$ with the conditioned $c$ directly. Instead it learns the distribution over each embedding and thereby learns a new embedding space. The samples of the learned space are passed on to generator with $z$. This approach is more discriminative in the latent space, and helps to separate two classes in the original generated space.

\begin{algorithm}[tb]
\caption{Training the proposed model} \label{algo:training}
\small
\begin{algorithmic}[1]
\For{$it$ in 1 \ldots n}
\State /* \textbf{Update P  (E-step)} */ 
\For{$j$ in 1 \ldots it} \hfill /* train GAN */
\For{$l$ in 1 \ldots 5}
\State $\operatorname{minimize} \mathcal{L}_{D}$
\EndFor
\State $\underset{\psi_{G}}{\operatorname{minimize}} \mathcal{L}_{G}$
\EndFor

\State /* \textbf{Update $\psi_{C}$  (M-step)} */
\For{$j$ in 1 \ldots it} \hfill /* train CSEM */
\State{Take text embedding $\varphi_{t_r}$ and $\varphi_{t_w} \sim  \varphi$}
\State{Obtain the $\hat{c_{tr}}$ and $\hat{c_{tw}}$ using the TE}
\State{Obtain $i_r$ and $i_{w}$ using Generator }
\State{$\underset{\psi_{C}}{\operatorname{minimize}} \mathcal{L}_{T}$}

\EndFor
\EndFor
\end{algorithmic}
\end{algorithm}

\subsubsection{\bf{Training setup and implementation details}} 
The values of the hyper-parameters in the model are set to $\alpha = 0.5$, $\beta = 2$,  and $\lambda = 2$ using grid-search.
The generator and the discriminator are trained in an iterative manner with the given objective functions (Eqns.~\ref{eq:D_new} and ~\ref{eq:G_new}). 
Both the generator and the discriminator are optimized with the root mean squared propagation(RmsProp) optimizer.  The generator is a fully connected neural network with two hidden layers. The first hidden layer has $2,048$ units and the second hidden layer has $4,096$ units. 
{\it Leaky ReLU} activation function has been used in the hidden layer, and the output of the generator is passed through a {\it ReLU} function to ensure that there are no negative values.
The discriminator is also a fully connected neural network with $1,024$ hidden layer units and 1 output unit. {\it Leaky ReLU} activation function is used in the hidden layer and no activation function is used for the output layer. 
The Text Encoder is a fully connected neural network with 2048 output units -- 1024 dimension for the mean ($\mu$) and 1024 dimension for the standard deviation ($\Sigma$).
The CSEM is implemented as a single layer feed forward network with {\it ReLU} activation. The weights and biases of both the generator and the discriminator are initialized with a random normal initializer having mean $0.0$ and standard deviation $0.02$.

\subsection{Applying model for retrieval}\label{sec:RetrievalSetup}

\begin{algorithm}[tb]
\small
\caption{Retrieval Algorithm}\label{algo:retrieval}
\begin{algorithmic}[1]
\Procedure{GetRelevantImages}{$\varphi_{t}$ , $k$}
\State\Comment{$\varphi_{t}$: the text embedding of a query-class}
\State\Comment{$k$: number of images to be retrieved}
\State{$\hat{c_t} \gets \Call{textEncoder}{\varphi_{t}}$}
\State{$z \gets \Call{getRandomNormalNoise}{}$}
\State{$i_t \gets \Call{G}{z, \hat{c_t}}$}
\State{$\theta_t \gets \Call{CSEM}{i_t}$}
\State{$simList \gets []$}
\For{$I_{us}$ in test\_image\_set}
\State{$I_i \gets \Call{FetchImgEmbedding}{I_{us}}$}
\State{$\theta_i \gets \Call{CSEM}{I_i}$}
\State{$sim_{it} \gets \Call{cosineSim}{\theta_t, \theta_i}$}
\State{$\Call{Append}{simList, <sim_{it}, I_{us}>}$}
\EndFor
\State \text{sort $simList$ in descending order of $sim_{it}$}
\State{$imageList \gets \text{images in first k indices of simList}$}
\State \textbf{return} $imageList$
\EndProcedure
\end{algorithmic}
\end{algorithm}

\noindent Once the model \model{} is trained, the retrieval of images for a novel/unseen class proceeds as shown in Algorithm~\ref{algo:retrieval}. 
The query for retrieval is the text embedding $\varphi_{t}$ of a novel/unseen class. 
Given $\varphi_{t}$, $\hat{c_t}$ is generated by the {\it text encoder}. 
Then $G()$ produces the image embedding $i_t$ which is passed through CSEM() to produce $\theta_t$. 
Now, for each image $I_{us}$ from an unseen class in the test set, we obtain the corresponding image embedding $I_i$
which is also passed through CSEM() to get $\theta_i$. 
Let $sim_{it} = cosineSim( CSEM(I_i), CSEM(G(z,\hat{c_t})) )$ be the cosine similarity between $\theta_t$ and $\theta_i$, 
where 
$z$ is a noise vector sampled from a random normal distribution.
Thereafter, a 2-tuple $<sim_{it}, I_{us}>$  is formed and appended to a list called $simList$. The list is then sorted in descending order of the $sim_{it}$ values. The top $k$ images are extracted from the sorted $simList$ and stored in $imageList$ which is returned as the ranked list of retrieved images.


\section{Experiments and Analysis} \label{sec:expt}

\noindent This section details our experiments through which we compare the performance of the proposed model with that of several state-of-the-art baselines, over several standard datasets. 

\subsection{Datasets} 
\label{sub:datasets}

\begin{table*}[tb]
    \footnotesize
    \center
    \addtolength{\tabcolsep}{2pt}
    \begin{tabular}{|c|p{1.2cm}|p{1cm}|p{1.5cm}|p{3cm}|p{3cm}|p{3cm}| }
        \hline
        \textbf{Dataset} & \textbf{Dimensions of T/A/I} & \textbf{\# T/A/I} & \# \textbf{Seen / Unseen Classes} & \textbf{Text Embedding Source} & \textbf{Image Embedding Source} & \textbf{Interpretation of Query}\\ 
        \hline
        \textbf{FLO} & T: 1024, \newline I: 2048 & T: 102, \newline I: 8,189 & 82/20  & \url{https://github.com/reedscot/cvpr2016}~\cite{scottreed-cvpr16} & \url{http://datasets.d2.mpi-inf.mpg.de/xian/ImageNet2011_res101_feature.zip}~\cite{xianCVPR17} & Textual description of a particular category of flowers\\
        \hline
        \textbf{CUB}  & T: 1024, \newline  I: 2048  & T: 200, \newline  I: 11,788 & 150/50 & \url{https://github.com/reedscot/cvpr2016}~\cite{scottreed-cvpr16} & \url{http://datasets.d2.mpi-inf.mpg.de/xian/ImageNet2011_res101_feature.zip}~\cite{xianCVPR17} & Textual description of a particular category of birds\\
        \hline
        \textbf{NAB} & T: 13217, \newline  I: 3072 & T: 404, \newline  I: 48,562 & 323/81 & \url{https://github.com/EthanZhu90/ZSL_GAN}~\cite{ZSL_GAN_cvpr2018} & \url{https://github.com/EthanZhu90/ZSL_GAN}~\cite{ZSL_GAN_cvpr2018} & Textual description of a particular category of birds\\
        \hline
        \textbf{AWA} & A: 85, \newline  I: 2048 & A: 50, \newline  I: 30,475 & 40/10 & \url{http://datasets.d2.mpi-inf.mpg.de/xian/ImageNet2011_res101_feature.zip}~\cite{xianCVPR17} & \url{http://datasets.d2.mpi-inf.mpg.de/xian/ImageNet2011_res101_feature.zip}~\cite{xianCVPR17} & Human annotated attributes of a particular category of animals\\
        \hline
        \textbf{Wiki} & T: 10, \newline  I: 128 & T: 2,866, \newline  I: 2,866 & 8/2 (following split in~\cite{CZHash-ICDM2020}) & \url{ttp://www.svcl.ucsd.edu/projects/crossmodal/}~\cite{wikidata-MM10} & \url{http://www.svcl.ucsd.edu/projects/crossmodal/}~\cite{wikidata-MM10} & Textual part of Wikipedia articles \\
        \hline    

    \end{tabular}
    \caption{{\bf Statistical description of the datasets (T: text, A: attribute, I: image). All models, including the proposed and baseline models, use the same embeddings for fair comparison (details in Section~\ref{sub:datasets}.}}
    \label{tab:dataset-description}
    \vspace{-8mm}
\end{table*}




We use the following five datasets for the experiments. Statistics of the datasets are summarized in Table~\ref{tab:dataset-description}.
For each dataset, the models work on image embeddings and text/attribute embeddings (e.g., of the image captions). 
Table~\ref{tab:dataset-description}
also states the sources for the various embeddings for each dataset.
For fair evaluation, every model, including the proposed and the baseline models, use the same text and image embeddings.

\vspace{1mm}
\noindent {\bf (1) Oxford Flowers-102 (Flowers) dataset} contains 8,189 images of flowers from 102 classes~\cite{Nilsback08}. 
Each class consists of between 40 and 258 images. Each image is accompanied by 10 sentences, each describing the image~\cite{scottreed-cvpr16}. 
The data is split into 82 training and 20 test classes, with {\it no overlap} among the training and test classes~\cite{Nilsback08}.

As text embeddings, we use charCNN-RNN embeddings of the image captions provided by~\cite{scottreed-cvpr16}.
We use the image embeddings provided by~\cite{xianCVPR17},
which are generated by passing each image through ResNet 101 architecture~\cite{resnet} pre-trained on the Imagenet~\cite{imagenet_cvpr09}, and taking the pre-trained ResNet's final layer embeddings.


\vspace{1mm}
\noindent {\bf (2) Caltech-UCSD Birds (CUB) dataset}, which contains 11,788 images of birds, divided into 200 species (classes), with each class containing approx. 60 images~\cite{WelinderEtal2010}. Each image has $10$~associated sentences describing the image~\cite{scottreed-cvpr16}. 
Following~\cite{xianCVPR17}, the images in  CUB are split into 150 training classes and 50 test classes, such that there is {\it no overlap} among the classes in training and test sets. 
Similar to the Flowers dataset, we use the text embeddings provided by~\cite{scottreed-cvpr16} and image embeddings provided by~\cite{xianCVPR17}.

\vspace{1mm}
\noindent {\bf (3) North American Birds (NAB) dataset} is a larger version of the CUB dataset, with 48,562 images of birds categorized into 1,011 classes.
The  dataset was extended by Elhoseiny et. al.~\cite{8100149} by adding a Wikipedia article for each class. The authors also merged some of the classes to finally have $404$ classes. 
Two types of split have been proposed in~\cite{8100149} in terms of how the seen (S) and unseen (U) classes are related -- 
(1)~Super-Category-Shared (SCS), and (2)~Super-Category-Exclusives (SCE). 
In SCS, unseen classes are chosen such that for each unseen class, there exists at least one seen class that have the {\it same parent category}. 
However, in SCE split, there is no overlap in the parent category among the seen and unseen classes.
Hence, retrieval is easier for the SCS split than for the SCE split. 

We use the text and image embeddings provided by~\cite{ZSL_GAN_cvpr2018}.
The $13,217$-dimensional text embeddings are actually TF-IDF vectors obtained from the Wikipedia articles (suitably preprocessed) corresponding to the classes.
The image embeddings are obtained by feeding the images into VPDE-net~\cite{VPDENet}, and extracting the activations of the part-based FC layer of the VPDE-net.
The NAB dataset has six semantic parts (`head', `back', `belly', `breast',  `wing', and `tail'). A 512-dimension feature vector is extracted for each part and concatenated in order. The resulting $3072$-dimensional vectors are considered as image embeddings.


\vspace{1mm}
\noindent {\bf (4) Animals with Attributes (AWA) dataset} consists of  $30,475$ images of animals from $50$ different classes. 
$85$ attributes are used to characterize each of the $50$ categories, giving an class-level attribute embedding of $85$ dimensions. 
The $2048$-dimension image embeddings are taken from the final layer ResNet-101 model trained on ImageNet~\cite{imagenet_cvpr09}.
There exists two kinds of splits of the data according to~\cite{xianCVPR17} -- { \it Standard-split} and {\it proposed split}. The standard-split does not take into consideration the ImageNet classes, while the proposed-split consists test classes which have no overlap with the ImageNet classes.
We used the proposed-split~\cite{xianCVPR17} of the dataset for the ZSIR models (including the proposed model, DADN, ZSL-GAN). 

\vspace{1mm}
\noindent {\bf (5) Wikipedia (Wiki) dataset} 
consists of $2,866$ image-text pairs taken from Wikipedia documents~\cite{wikidata-MM10}. The image-text pairs have been categorized into $10$ semantic labels (classes). 
Images are represented by $128$-dimensional SIFT feature vectors while the textual data are represented as probability distributions over the $10$ topics, which are derived from a Latent Dirichlet Allocation (LDA) model. Following~\cite{CZHash-ICDM2020}, the classes are split randomly in an 80:20 train:test ratio, and average of results over $10$ such random splits is reported.

\vspace{1mm}
\noindent {\bf Ensuring zero-shot setup for some of the datasets:} The Flowers, CUB, and AWA datasets use image embeddings from ResNet-101 that is pretrained over ImageNet. 
Hence, for these datasets, 
it is important to ensure that the test/unseen classes should {\it not} have any overlap with the ImageNet classes, otherwise such overlap will violate the ZSL setting~\cite{xianCVPR17}. 
Therefore, for all these three datasets, we use the same train-test class split as proposed by~\cite{xianCVPR17}, which ensures that there is {\it no overlap among the test classes and the ImageNet classes on which ResNet-101 is pretrained}.

\vspace{-2mm}
\subsection{Evaluation Metrics} 

For each class in the test set, we consider as the {\it query} its per-class text embedding $\varphi_{t}$ (or the per-class attribute embedding in case of the AWA dataset). 
The physical interpretation of the query for each dataset is explained in Table~\ref{tab:dataset-description} (last column). 
For each class, we retrieve $50$ top-ranked images (according to each model).
Let the number of queries (test classes) be $Q$.
We report the following evaluation measures:

\noindent {\bf (1) Precision@50, averaged over all queries:}
For a certain query $q$, Precision@50($q$) = $\frac{k}{50}$, where $k$ is the number of relevant images among the top-ranked $50$ images retrieved for $q$. 
We report Precision@50 averaged over all $Q$ queries.


\noindent {\bf (2) mean Average Precision (mAP@50):}
mAP@50 is the mean of {\it Average Precision at rank $50$}, where the mean is taken over all $Q$ queries. $mAP@50 = \frac{1}{Q}\sum_{q=1}^Q \operatorname{AveP_{50}(q)}$
where 
$\operatorname{AveP_{50}(q)}$ for query $q$ is $\frac{\sum_{r \in R} \operatorname{Precision@r(q)}}{|R|}$
where $R$ is the set of ranks (in $[1, 50]$) at which a relevant image has been found for $q$.

\noindent {\bf (2) Top-1 Accuracy (Top-1 Acc):} 
This metric measures the fraction of queries (unseen classes) for which the {\it top-ranked} retrieval result is relevant~\cite{ZSL_GAN_cvpr2018,scottreed-cvpr16}.
In our experiments we report the average Top-1 Accuracy of the models over the set of all unseen classes.

\vspace{-2mm}
\subsection{Baselines} \label{sub:baselines}

We compare the proposed \model{} model with different kinds of baseline models, as described below.

\vspace{2mm}
\noindent {\bf Zero-Shot Information Retrieval (ZSIR) models:}
We consider three state-of-the-art models for ZSIR: 

\noindent (1) Reed {\it et al.} developed the {\bf DS-SJE~\cite{scottreed-cvpr16}} model that jointly learns the image and text embeddings using a joint embedding loss function. 
We use the codes and pre-trained models provided by the authors (at \url{https://github.com/reedscot/cvpr2016}). 

\noindent (2) Chi {\it et al.} proposed two very similar models for zero-shot IR~ \cite{8643797,ijcai2018-92}. 
The {\bf DADN model~\cite{8643797}} out-performs the one in~\cite{ijcai2018-92}.
Hence, we consider DADN as our baseline. DADN uses dual GANs to exploit the category label embeddings to project the image embeddings and text embeddings to have a common representation in a semantic space~\cite{8643797}. 
We use the codes provided by the authors (at \url{https://github.com/PKU-ICST-MIPL/DADN_TCSVT2019}).

\noindent (3) {\it Image generation models} are  optimized to generate high-fidelty images from given textual descriptions. 
These models can be used for text-to-image retrieval as follows -- given the textual query, we can use such a model to generate an image, and then retrieve images that are `similar' to the generated image as answers to the query. 
Zhu {\it et al.}~\cite{ZSL_GAN_cvpr2018} developed such a GAN-based approach for retrieval of images from textual data. The GAN (called {\bf ZSL-GAN}) is trained with classification loss as a regularizer.
We use the codes provided by the authors (at \url{https://github.com/EthanZhu90/ZSL_GAN}).

\vspace{2mm}
\noindent {\bf Zero-Shot Classification (ZSC) models:}
We consider two state-of-the-art ZSC models~\cite{verma-cvpr18,xian2018feature_gen} (both of which use generative models), and adopt them for the retrieval task as described below. 

\noindent (1) {\bf f-CLSWGAN~\cite{xian2018feature_gen}}
used the GAN to synthesize the unseen class samples (image embeddings) using the class attribute. 
Then, using the synthesized samples of the unseen class, they trained the softmax classifier. We used the codes provided by the authors (at  \url{http://datasets.d2.mpi-inf.mpg.de/xian/cvpr18xian.zip}).
To use this model for retrieval, we used the image embeddings generated by the generator for the unseen classes, and ranked the image embeddings of the unseen classes using their cosine similarity with the generated image embedding.

\noindent (2) {\bf SE-ZSL~\cite{verma-cvpr18}} used the Conditional-VAE (CVAE) with feedback connection for Zero-Shot Learning (implementation obtained on request from the authors).
We adopt the same architecture for the text to image retrieval as follows. Using the CVAE, we first trained the model over the training class samples. At test time, we generated the image embedding of the unseen classes conditioned on the unseen text query. In the image embedding space, we perform the nearest neighbour search between the generated image embedding and the original image embedding.


\vspace{2mm}
\noindent {\bf Zero-Shot Hashing models:} Hashing based retrieval models have been widely studied because of their low storage cost and fast query speed. We compare the proposed model with some Zero-Shot Hashing models, namely DCMH~\cite{DCMH-AAAI17}, SePH~\cite{SePH-ITC17}, and the more recent AgNet~\cite{AgNet-TNNLS2019} and CZHash~\cite{CZHash-ICDM2020}.
The implementations of most of these hashing models are not available publicly; hence we adopt the following approach. 
We apply our method to the same datasets (e.g., AwA, Wiki) for which results have been reported in the corresponding papers~\cite{AgNet-TNNLS2019, CZHash-ICDM2020}, and use the same experimental setting as reported in those papers.

\vspace{2mm}
\noindent Note that different prior works have reported different performance metrics. For those baselines whose implementations are available to us, we have changed the codes of the baselines minimally, to report Precision@50, mAP@50, and Top-1 Accuracy for all models.
For comparing with the hashing models (whose implementations are not available), we report only mAP which is the only metric reported in~\cite{AgNet-TNNLS2019, CZHash-ICDM2020}.

\begin{table}[tb]
    \footnotesize
    \center
    \addtolength{\tabcolsep}{-0pt}
    \begin{tabular}{|l|c|c|c|}
        \hline
        \textbf{Retrieval Model} & \textbf{Prec@50(\%)} & \textbf{mAP@50(\%)} & \textbf{Top-1 Acc(\%)} \\ 
        \hline
        \multicolumn{4}{|c|}{Zero-shot classification models adopted for retrieval} \\ \hline
        SE-ZSL \cite{verma-cvpr18} & 29.3\%& 45.6\% & 59.6\% \\
        fCLSWGAN \cite{xian2018feature_gen} &36.1\% & 52.3\% &64\% \\ \hline
        \multicolumn{4}{|c|}{Zero-shot retrieval models} \\ \hline        
        DS-SJE~\cite{scottreed-cvpr16} & 45.6\% &58.8\% &54\% \\
        ZSL-GAN~\cite{ZSL_GAN_cvpr2018} & 42.2\% &59.2\% &60\% \\
        DADN~\cite{8643797} &48.9\% & 62.7\% &68\% \\
        \hline    
        {\bf \model{} (proposed)} & {\bf 52}\%$^{SFJGD}$  & {\bf 65.4}\%$^{SFJGD}$  & {\bf 74}\%$^{SFJGD}$ \\
        \hline
    \end{tabular}
        \caption{{\bf Zero-Shot Retrieval on CUB dataset. The proposed model outperforms all  baselines (bold-font indicates the best results in all tables). The super-scripts S, F, J, G, and D indicate that the proposed method is statistically significantly better at 95\% confidence level (p < 0.05) than SE-ZSL, fCLSWGAN, DS-SJE, ZSL-GAN and DADN respectively.}}
    \label{tab:result-cub}
    \vspace{-6mm}
\end{table}

\begin{table}[tb]
    \footnotesize
    \center
    \addtolength{\tabcolsep}{-0pt}
    \begin{tabular}{| l | c | c | c |}
        \hline
        \textbf{Retrieval Model} & \textbf{Prec@50(\%)}  & \textbf{mAP@50(\%)} & \textbf{Top-1 Acc(\%)}  \\ 
        \hline
        \multicolumn{4}{|c|}{Zero-shot classification models adopted for retrieval} \\ \hline
        SE-ZSL~\cite{verma-cvpr18} & 41.7\% & 63.1\% & 66.4\%\\
        fCLSWGAN~\cite{xian2018feature_gen} & 44.1\% & 67.2\% &71.2\%\\
        \hline
        \multicolumn{4}{|c|}{Zero-shot retrieval models} \\ \hline        
        DS-SJE~\cite{scottreed-cvpr16} & 55.1\% & 65.7\% &63.7\%\\
        ZSL-GAN~\cite{ZSL_GAN_cvpr2018} & 38.7\% &46.6\% &45\%\\
        DADN~\cite{8643797} &20.8\% & 28.6\%&25\%\\
        \hline    
        {\bf \model{} (proposed)} & {\bf 59.5}\%$^{SFJGD}$ & {\bf 69.4}\%$^{SFJGD}$ &{\bf 80}\%$^{SFJGD}$ \\
        \hline
    \end{tabular}
    \caption{{\bf Zero-Shot Retrieval on Flower dataset. The proposed model outperforms all the baselines. Superscripts S, F, J, G, D show significant improvements (see Table~\ref{tab:result-cub}).}}
    \label{tab:result-flower}
    \vspace{-8mm}
\end{table}

\subsection{Comparing performances of models}

Table~\ref{tab:result-cub} and Table~\ref{tab:result-flower} respectively compare the performance of the various models on the CUB and Flower datasets.
Similarly, Table~\ref{tab:result-nab-hard} and Table~\ref{tab:result-nab-easy} compare performances over the NAB dataset (SCE and SCS splits respectively).
Table~\ref{tab:result-AWA2-Wiki-retrieval} compares performances over the AwA and Wiki datasets. The main purpose of Table~\ref{tab:result-AWA2-Wiki-retrieval} is to compare performance with the Zero-Shot Hashing models whose implementations are not available to us; hence we report only mAP which is the only metric reported in~\cite{AgNet-TNNLS2019, CZHash-ICDM2020}. 

The proposed \model{} considerably outperforms almost all the baselines across all datasets, the only exception being that DADN outperforms \model{} on the Wiki dataset (Table~\ref{tab:result-AWA2-Wiki-retrieval}). 
We performed Wilcoxon signed-rank statistical significance test at a confidence level of 95\%. The superscripts S, F, J, G, and D in the tables indicate that the proposed method is statistically significantly better at 95\% confidence level ($p < 0.05$) than SE-ZSL, fCLSWGAN, DS-SJE, ZSL-GAN and DADN respectively.
We find that the results of the proposed model are significantly better than most of the baselines. 
Note that we could not perform significance tests for the Hashing methods owing to the unavailability of their implementations.


We now perform a detailed analysis of why our model performs better than the baselines.


\begin{table}[tb]
    \footnotesize
    \center
    \addtolength{\tabcolsep}{-0pt}
    \begin{tabular}{| l | c | c | c |}
        \hline
        \textbf{Retrieval Model} & \textbf{Prec@50(\%)}  & \textbf{mAP@50(\%)} & \textbf{Top-1 Acc(\%)} \\ 
        \hline
        \multicolumn{4}{|c|}{Zero-shot classification models adopted for retrieval} \\ \hline
        SE-ZSL~\cite{verma-cvpr18} &7.5\% & 3.6\% &7.2\%\\
        \hline
        \multicolumn{4}{|c|}{Zero-shot retrieval models} \\ \hline        
        ZSL-GAN~\cite{ZSL_GAN_cvpr2018} & 6\% &9.3\% &6.2\%\\
        DADN~\cite{8643797} &4.7\% & 7.3\% &2.5\%\\
        \hline    
        {\bf \model{} (proposed)} & {\bf 8.4}\%$^{GD}$ & {\bf 11.8}\%$^{SGD}$ &{\bf 7.4}\%$^{GD}$\\
        \hline
    \end{tabular}
    \caption{{\bf Zero-Shot Retrieval on NAB dataset (SCE split). The proposed model outperforms all the baselines. Superscripts S, F, J, G, D show significant improvements (see Table~\ref{tab:result-cub}).}}
    \label{tab:result-nab-hard}
    \vspace{-8mm}
\end{table}

\vspace{1mm}
\noindent {\bf \model{} vs. DS-SJE:}
DS-SJE~\cite{scottreed-cvpr16} uses discriminative models to create text and image embeddings.\footnote{We could not run DS-SJE over some of the datasets, as we were unable to modify the Lua implementation to suit these datasets. Also, the Prec@50 of DS-SJE on the Flower dataset (in Table~\ref{tab:result-flower}) is what we obtained by running the pre-trained model provided by the authors, and is slightly different from what is reported in the original paper.}
This discriminative approach has limited visual imaginative capability, whereas the generative approach of the proposed \model{} does not suffer from this problem.
Specifically, the low performance of DS-SJE on the CUB datast (Table~\ref{tab:result-cub}) is due to
the lack of good visual imaginative capability which is required to capture the different postures of birds in the CUB dataset (which is not so much necessary for flowers).

\begin{table}[tb]
    \footnotesize
    \center
    \addtolength{\tabcolsep}{-0pt}
    \begin{tabular}{| l | c | c | c |}
        \hline
        \textbf{Retrieval Model} & \textbf{Prec@50(\%)}  & \textbf{mAP@50(\%)} & \textbf{Top-1 Acc(\%)}  \\ 
        \hline
        \multicolumn{4}{|c|}{Zero-shot classification models adopted for retrieval} \\ \hline
        SE-ZSL~\cite{verma-cvpr18} & 25.3\% & 34.7\% &11.4\%\\
        \hline
        \multicolumn{4}{|c|}{Zero-shot retrieval models} \\ \hline        
        ZSL-GAN~\cite{ZSL_GAN_cvpr2018} & 32.6\% & 39.4\% &34.6\%\\
        DADN~\cite{8643797} &26.5\% & 28.6\% &17.3\%\\
        \hline
        {\bf \model{} (proposed)} & {\bf 36}\%$^{SGD}$ & {\bf 43}\%$^{SGD}$ &{\bf 49.4}\%$^{SGD}$ \\
        \hline
    \end{tabular}
    \caption{{\bf Zero-Shot Retrieval on NAB dataset on (SCS split). The proposed model outperforms all the baselines. Superscripts S, G, D show significant improvements (see Table~\ref{tab:result-cub}).}}
    \label{tab:result-nab-easy}
    \vspace{-8mm}
\end{table}


\begin{table}[tb]
    \footnotesize
    \center
    \addtolength{\tabcolsep}{-3.5pt}
    \begin{tabular}{| l | c | c | }
        \hline
        \textbf{Retrieval Model} & \textbf{mAP(\%) on AwA} &  \textbf{mAP(\%) on Wiki} \\ 
        \hline
        \multicolumn{3}{|c|}{Zero-shot Hashing models} \\ \hline        
        DCMH~\cite{DCMH-AAAI17} & 10.3\% (with 64-bit hash) &  24.83\% (with 128-bit hash)
	    \\   
        AgNet~\cite{AgNet-TNNLS2019} & 58.8\% (with 64-bit hash) & 25.11\% (with 128-bit hash)
	    \\   
    	SePH~\cite{SePH-ITC17} & -- &  50.44\% (with 128-bit hash)
	    \\  
    	CZHash~\cite{CZHash-ICDM2020} &  --  &  25.87\% (with 128-bit hash)
	    \\   \hline
        \multicolumn{3}{|c|}{Zero-shot Retrieval models} \\ \hline	    
        ZSL-GAN~\cite{ZSL_GAN_cvpr2018} & 12.5\%  & -
	    \\ 
        DADN~\cite{8643797} & 27.9\%  &  {\bf 58.94\%} 
	    \\ \hline
        {\bf \model{} (proposed)} & {\bf 62.2}\%$^{GD}$ & 56.9\%
	    \\   \hline
    \end{tabular}
    \caption{{\bf Zero-shot retrieval on (i)~AwA dataset, and (ii)~Wiki dataset. Results of hashing models  reproduced from~\cite{AgNet-TNNLS2019} and~\cite{CZHash-ICDM2020}. Other 
    metrics could not be reported due to unavailability of the implementations of the hashing models.}} 
    \label{tab:result-AWA2-Wiki-retrieval}
    \vspace{-8mm}
\end{table}

\vspace{1mm}
\noindent {\bf \model{} vs. DADN:}
DADN~\cite{8643797} uses semantic information contained in the class labels to train the GANs to project the textual embeddings and image embeddings in a common semantic embedding space. 
Specifically, they use 300-dimensional Word2vec embeddings pretrained on the Google News corpus\footnote{\url{https://code.google.com/archive/p/word2vec/}}, to get embeddings of the class labels. 
A limitation of DADN is that unavailability of proper class label embeddings can make the model perform very poorly in retrieval tasks. 
For instance, DADN performs extremely poorly on the Flowers dataset, 
since out of the $102$ class labels in the Flowers dataset, the pretrained Word2vec embeddings are not available for as many as $23$ labels. 
Similarly, out of $404$ class labels in the NAB dataset, the pretrained Word2vec embeddings are not available for $8$ labels.
\model{} does {\it not} rely on class labels, and performs better than DADN on both these datasets. 
On the other hand, DADN performs better than \model{} on the Wiki dataset (Table~\ref{tab:result-AWA2-Wiki-retrieval} since pretrained embeddings are available for all class labels.

\vspace{1mm}
\noindent {\bf \model{} vs. ZSL-GAN:}  ZSL-GAN uses a GAN to generate image embeddings from textual descriptions.\footnote{We could not run ZSL-GAN on the Wiki dataset, since this model assumes a single text description $T_c$ for each class $c$, whereas each class has more than one textual description in the Wiki dataset.} 
In their model the discriminator branches to two fully connected layers -- one for distinguishing the real image embeddings from the fake ones, and the other for classifying real and fake embeddings to the correct class. 
We believe that the retrieval efficiency is lower due to  this classification layer of the discriminator. This layer learns to classify the real and generated embeddings to the same class, which is contradictory to its primary task of distinguishing between the real and generated embeddings (where ideally it should learn to assign different labels to the real and generated embeddings).

\vspace{1mm}
\noindent {\bf \model{} vs. Hashing models (e.g., AgNet, CZHash):}
Hashing-based retrieval methods are popular due to their low storage cost and fast retrieval. However, achieving these comes at a cost of retrieval accuracy -- to generate the hash, the  models sacrifice some information content of the embeddings, and this results in the loss in accuracy. 
As can been seen from Table ~\ref{tab:result-AWA2-Wiki-retrieval}, all the hashing methods have substantially lower mAP scores than our proposed method.




\subsection{Analysing two design choices in \model{}}\label{sec:analysis}

In this section, we analyse two design choices in our proposed model --
(1)~why we adopted an Expectation-Maximization approach, instead of jointly training the GAN and the CSEM, and 
(2)~why we chose to select wrong class embeddings {\it randomly}.

\subsubsection{\bf E-M vs Joint Optimization:}
\begin{figure}[tb]
\centering
\includegraphics[width=0.8\columnwidth, height=3cm]{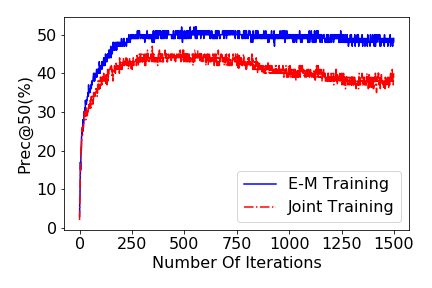}
\vspace{-5mm}
\caption{{\bf [color online] Comparing performance of proposed E-M setup (solid blue curve) with joint training of the GAN and the CSEM (dashed red curve). Shown is how Prec@50 varies with the number of iterations for which the model is trained, over the CUB dataset. The E-M setup achieves consistently better performance.}}
\label{fig:em-vs-joint}
\vspace{-5mm}
\end{figure}

In the proposed model, the CSEM and the GAN are trained alternately using an E-M setup (see Sec.~\ref{sec:EMsetup}). 
However, it is also possible to train both these networks jointly; i.e., when the Generator is trained, the CSEM loss is also optimized.
We performed a comparison between these two approaches, and observed 
that the performance of the model drops by a significant amount when jointly trained. 
Figure~\ref{fig:em-vs-joint} compares the performance of the two approaches in terms of Precision@50 over the CUB dataset, and shows that the EM setup results in better retrieval.
Similar observations were made for other performance metrics, across all datasets (details omitted for brevity).

The reason why the jointly training approach, where the CSEM loss is  optimized along with the generator loss, does not work well is as follows.
Using the triplet Loss $\mathcal{L}_{T}$ (defined in Eqn.~\ref{eq:TripletLoss}), the CSEM learns 
maximize similarity with the relevant embeddings and minimize similarity with the irrelevant embeddings.
Thus, during backpropagation, the relevant representative embeddings have a different gradient than the irrelevant representative embeddings $G(z,\hat{c_{tw}})$. 
When the CSEM is jointly trained with the generator, the weights of the CSEM get updated after every iteration, causing different gradients (or weights) for each of the wrong class embeddings of the generator, thus causing a distorted space of wrong class embeddings, and thus causing hindrance to the learning of the Generator. This problem, however, has been removed in the EM setup where the parameters of the generator is frozen while training the CSEM.  


\begin{figure}[t]
\centering
\includegraphics[width=0.49\columnwidth]{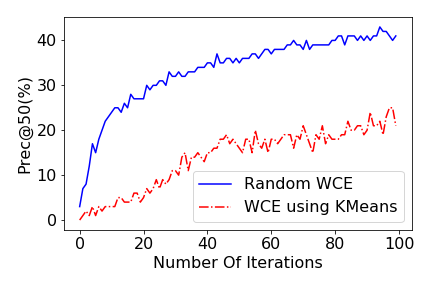}
\includegraphics[width=0.49\columnwidth]{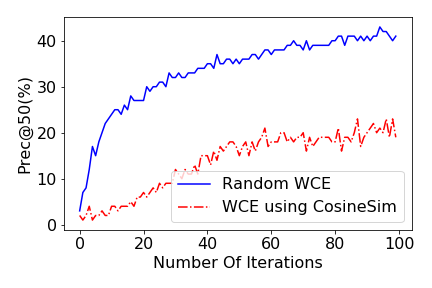}
\vspace{-5mm}
\caption{{\bf [color online] Comparing random selection of Wrong Class Embeddings (WCE) with other ways of selecting WCE: (a)~using KMeans clustering of images, and (b)~using cosine similarity (details in text). Both figures show how Prec@50 varies with number of iterations for which the model is trained, over the CUB dataset. Random selection of WCE performs the best.}}
\label{fig:wrong-acc}
\vspace{-8mm}
\end{figure}


\subsubsection{\bf Choice of wrong class embeddings:}

In the proposed model, for given $I_r$ and $\phi_{tr}$ for a certain target class $c_{target}$, we learn the representative embedding for the images of that class. To this end, we use wrong class embeddings (WCE) selected randomly from among all other classes. 
One might think that, instead of randomly selecting wrong classes, we should employ some intelligent strategy of selecting wrong classes. For instance, one can think of selecting WCE such that they are most similar to $\phi_{tr}$, which may have the apparent benefit that the model will learn to distinguish between confusing (similar) classes well.
However, we argue otherwise.

Restricting the choice of the wrong classes distorts the space of the wrong embeddings, and hence runs the risk of the model identifying embeddings from classes outside the space of distorted wrong embeddings as relevant. 
In other words, though the model can learn to distinguish the selected wrong classes well, it fails to identify other classes (that are {\it not} selected as the wrong classes) as non-relevant. 

To support our claim, we perform two experiments -- 
(1)~The wrong class is selected as the class whose text embedding has the highest cosine similarity to $\phi_{tr}$, and 
(2)~The images are clustered using K-Means clustering algorithm, and the wrong class is selected as that class whose images co-occur with the maximum frequency in the same clusters as the images from $c_{target}$. 
Figure~\ref{fig:wrong-acc} compares the performance of the proposed model (where WCE are selected randomly) with these two modified models. Specifically Precision@50 is compared over the CUB dataset. In both cases, the accuracy drops drastically when WCE are chosen in some way other than randomly. Observations are similar for other performance metrics and other datasets (omitted for brevity).

\subsection{Error Analysis of \model{}}

\begin{figure}[t]
\centering
\includegraphics[width=0.95\columnwidth]{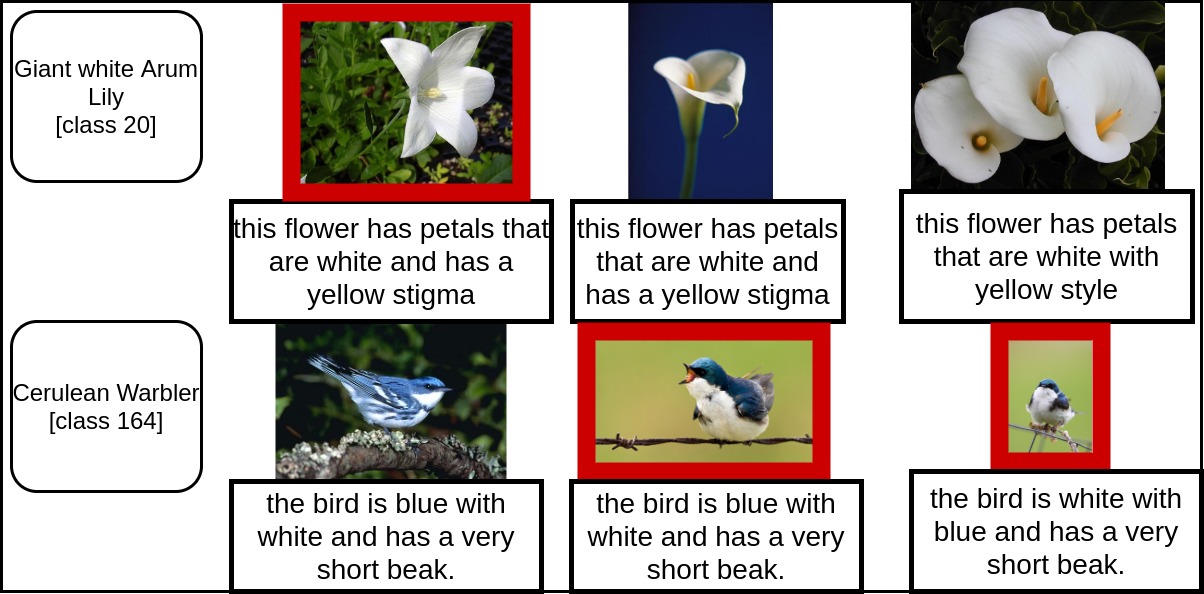}  
\vspace{-5pt}
\caption{{\bf [color online] Images from the top three classes retrieved by \model{}, for the query-classes shown on the left. 
Top panel for Flower dataset, bottom panel for CUB dataset. 
The images with thick red boundaries are from some class that is {\it not} the query-class (hence not considered relevant), but they are very similar to some images in the query-class.}}
\label{fig:failure-cases}
\vspace{-4mm}
\end{figure}

We analyse the failure cases where \model{} retrieves an image from a different class, compared to the query-class (whose text embedding has been issued as the query).
Figure~\ref{fig:failure-cases} shows some such examples, where the images enclosed in thick red boxes are not from the query-class.
In general, we find that the wrongly retrieved images are in fact very similar to some (correctly retrieved) images in the query-class.
For instance, in the CUB dataset, for the query-class {\it Cerulean Warbler}, the textual description of an image from this class ({\it this bird is blue with white and has a very short beak}) matches exactly with that of an image from a different class (which was retrieved by the model). 
Other cases can be observed where an image of some different class that has been retrieved,  matches almost exactly with the description of the query-class. 
For instance, in the Flower dataset, for the query class {\it Giant White Arum Lily}, the wrongly retrieved class also has flowers with white petals and a yellow stigma, which matches exactly with many of the flowers in the {\it Giant White Arum Lily} class. 


\subsection{Ablation Analysis of \model{}} \label{sec:ablation}

Table~\ref{tab:ablation_result} reports an ablation analysis, meant to 
analyze the importance of different components of our proposed architecture.
For brevity, we report only Prec@50 for the two datasets CUB and Flowers (observations on other datasets are qualitatively similar).

The largest drop in performance occurs when the wrong class embedding is not used.
As stated earlier, this use of wrong class embeddings is one of our major contributions, and an important factor in the model's performance.
Another crucial factor is the generation of representative embeddding $i \in I'$ for each class using a GAN. Removing this step also causes significant drop in performance.
The Triplet loss and the Log of odds ratio regularizer (R) are also crucial -- removal of either leads to significant degradation in performance.
Especially, if R is removed, the performance drop is very high for the Flower dataset. 
R is more important for the Flower dataset, since it is common to find different flowers having similar shape but different colors, and R helps to distinguish flowers based on their colors.

\begin{table}[!t]
    \footnotesize
    \center
    \addtolength{\tabcolsep}{-1pt}
    \begin{tabular}{|p{5.5cm}|c|c|}
        \hline
        \textbf{Retrieval Model} & \textbf{Prec@50} & \textbf{Prec@50} \\ 
          & {\bf CUB} & {\bf Flowers} \\ \hline
        Complete proposed model & 52\%  &  59.5\% \\ 
        \hline \hline
        w/o use of wrong class embedding  & 24.7\% & 27.2\% \\
        w/o R (regularizer) and Triplet Loss (CSEM)  & 23.8\% & 33.7\%  \\
        w/o Triplet Loss  & 36.2\% & 41.4\% \\
        w/o R (regularizer)   & 48.4\% & 35.2\% \\
        w/o GAN (i.e. the representative embedding for a class)   & 25.9\% & 32\% \\
        \hline
    \end{tabular}
        \caption{{\bf Results of ablation analysis on the proposed model. Precision@50 reported on CUB and Flower datasets.}}
        \vspace{-10mm}
    \label{tab:ablation_result}
\end{table}


\section{Conclusion}

We propose a novel model for zero-shot text to image (T $\rightarrow$ I) retrieval, which outperforms many state-of-the-art models for ZSIR as well as several ZS classification and hashing models on several standard datasets. 
The main points of novelty of the proposed model \model{} are 
(i)~use of an E-M setup in training, and (ii)~use of wrong class embeddings to learn the representation of classes. 
The implementation of \model{} is publicly available at \url{https://github.com/ranarag/ZSCRGAN}.

In future, we look to apply the proposed model to other types of cross-modal retrieval (I $\rightarrow$ T), as well as to the zero -shot multi-view setup (TI $\rightarrow$ I, TI $\rightarrow$ T, I $\rightarrow$ TI, etc.) where multiple modes can be queried or retrieved together. 

\vspace{3mm}
\noindent {\bf Acknowledgements:} The authors acknowledge the support of NVIDIA Corporation with the donation of the Titan Xp GPU used for this research.


\end{document}